\pdfoutput=1

\documentclass[conference,compsoc]{IEEEtran}
\usepackage{fancyhdr}
\ifCLASSOPTIONcompsoc
  \usepackage[nocompress]{cite}
\else
  \usepackage{cite}
\fi
\ifCLASSINFOpdf
  \usepackage[pdftex]{graphicx}
  \graphicspath{{./figures/}}
  \DeclareGraphicsExtensions{.pdf,.png}
\else
\fi
\usepackage{amsmath}
\begin{document}
%
\title{Gone but Not Forgotten: Improved Benchmarks for Machine Unlearning}

\author{
\IEEEauthorblockN{Keltin Grimes, Collin Abidi, Cole Frank, and Shannon Gallagher}
\IEEEauthorblockA{Carnegie Mellon University \\
Software Engineering Institute \\
Pittsburgh, Pennsylvania 15213 \\
\{kgrimes, cabidi, cfrank, skgallagher\}@sei.cmu.edu
}
}


\maketitle

\fancyhead[C]{\textit{Accepted as an Extended Abstract at the Deep Learning Security and Privacy Workshop 2024, co-located with IEEE S\&P}}
\thispagestyle{fancy}

\begin{abstract}
Machine learning models are vulnerable to adversarial attacks, including attacks that leak information about the model's training data. There has recently been an increase in interest about how to best address privacy concerns, especially in the presence of data-removal requests. Machine unlearning algorithms aim to efficiently update trained models to comply with data deletion requests while maintaining performance and without having to resort to retraining the model from scratch, a costly endeavor. Several algorithms in the machine unlearning literature demonstrate some level of privacy gains, but they are often evaluated only on rudimentary membership inference attacks, which do not represent realistic threats. In this paper we describe and propose alternative evaluation methods for three key shortcomings in the current evaluation of unlearning algorithms. We show the utility of our alternative evaluations via a series of experiments of state-of-the-art unlearning algorithms on different computer vision datasets, presenting a more detailed picture of the state of the field.
\end{abstract}


%
\IEEEpeerreviewmaketitle

\section{Introduction}

While incorporating new data into established machine learning models via fine-tuning is a well-studied problem, the inverse problem of removing data from those models has received less attention until recently. The field of \textit{machine unlearning} \cite{xu2023machine} \cite{nguyen2022survey} \cite{guo2019certified} has emerged over the last several years in response to legal frameworks, such as the EU's GDPR \cite{gdpr}, that give citizens the ``right to be forgotten.'' To comply with such legal frameworks, companies may be required not only to remove user data from data structures, but also from machine learning models trained on that data. Machine unlearning can be applied to many data domains including images, videos, audio, and text -- in this paper we focus on computer vision (CV) in response to recent research efforts to comply with legal requirements for user privacy.

The na\"ive approach to unlearning would be to erase the required data from a dataset and retrain a model from scratch. The removed data will have no influence on the model parameters or its output (ignoring indirect influences such as hyper-parameters optimized prior to removal), so an attacker would have no way of extracting information about the removed data. The major downside of this approach - and the main issue that the machine unlearning literature seeks to address - is the computational burden and hence the immense cost in terms of processes and technology. Retraining a model from scratch on unlearning requests that arrive in a stream may simply not be feasible at scale, especially for large datasets and models with hundreds of millions of parameters that require thousands of epochs on tens or hundreds of GPUs to train. Ideal unlearning algorithms would provide the privacy guarantees and performance characteristics of a retrained-from-scratch model and require far fewer compute resources. As such, machine unlearning involves balancing \textit{privacy}, \textit{cost}, and \textit{long-term performance}.

To realistically represent conditions under practical attacks when benchmarking machine unlearning algorithms, we argue that researchers and practitioners should consider the following three essential characteristics:
\begin{enumerate}
    \item Emphasis of \textit{worst-case} metrics over average-case metrics and the use of strong adversarial attacks to provide a high-quality upper-bound on privacy.
    \item Consideration of model \textit{update}-based attacks (e.g. \textit{leakage}), as in \cite{chen2021machine}, which may cause unlearning to provide \textit{additional} information to attackers.
    \item Analysis of unlearning algorithm performance over repeated applications of unlearning (i.e. \textit{iterative unlearning}), especially in regards to degradation of test accuracy performance of the unlearned models.
\end{enumerate}

In response to these gaps, we propose a framework that incorporates a suite of improved benchmarks for the testing and evaluation of machine unlearning algorithms. To test our framework and demonstrate the utility of the improved benchmarks we conduct a benchmarking study of a variety of state-of-the-art (SoTA) unlearning algorithms with the goal of presenting a holistic assessment of unlearning algorithms. Our contributions are intended to demonstrate the importance of using more comprehensive evaluations.

\subsection{Unlearning Definition and Notation}

We are primarily interested in \textit{approximate unlearning} where the influence of the forget set is practically removed, as opposed to exact unlearning where the influence must completely and provably removed. We define unlearning as follows. We have a dataset $\mathcal{D}$ partitioned into forget, retain, validation, and test sets $\mathcal{D}_f$, $\mathcal{D}_r$, $\mathcal{D}_{val}$, and $\mathcal{D}_{test}$, respectively. A model $M$ is trained on the training subset $\mathcal{D}_{train} = \mathcal{D}_{f} \cup \mathcal{D}_{r}$, where $\mathcal{D}_{f} \cap \mathcal{D}_{r} = \emptyset$. An unlearning algorithm $\mathcal{U} : M \times (\mathcal{D}_{r}, \mathcal{D}_{f}) \rightarrow M'$ produces a new model $M'$ which ideally has minimized the possibility of information leakage from $\mathcal{D}_{f}$, maintained model performance, and done so with minimal compute. Iterative unlearning requires an extended definition (omitted here for brevity), but intuitively, it is the repeated act of unlearning over time.

\fancyhead{}

\section{Evaluating Unlearning Algorithms}

Prior works in unlearning \cite{chen2021machine, foster2023fast, huang2023tight, graves2021amnesiac, ginart2019making, gupta2021adaptive, neel2020descenttodelete, zanella2020analyzing, foster2024zero} take inspiration from the Differential Privacy (DP) literature, using membership inference attacks (MIAs), which can determine whether a certain example was part of the training set, to demonstrate an empirical upper bound on privacy. Intuitively, such privacy auditing attacks demonstrate what is possible by an adversary, and thus high-confidence upper bounds are crucial for understanding vulnerabilities in a system.

\subsection{Privacy as a Worst-Case Metric} \label{sec:privacy-worst-case-metric}

We argue that worst-case measures of privacy are crucial for effective evaluations of unlearning algorithms. Users, in the absence of a strict guarantee of their own privacy, will care primarily about worst-case outcomes for themselves, rather than average- or even best-case metrics. When examining privacy through the lens of DP, as many in the unlearning field do, this becomes even more clear. Satisfying $(\epsilon, \delta)$-Differential Privacy requires a mathematical proof showing an algorithm satisfies a strict worst-case upper-bound on potential information leakage from $\mathcal{D}_{train}$ across all possible training examples. Therefore, strong unlearning evaluations will both use as effective MIAs as possible, and present the results of the attacks with worst-case metrics.   

Through our literature review we have found that the most common MIA used to evaluate algorithms is a simple Logistic Regression classifier trained to predict whether or not a loss value comes from an example in $\mathcal{D}_f$ \cite{foster2023fast, chundawat2023can, golatkar2020forgetting, kurmanji2023towards, choi2023towards}. Furthermore, results are reported almost exclusively through either accuracy or recall - both of which are average case metrics. The study in \cite{kurmanji2023towards} takes an important step forward in using stronger MIAs, adapting the online version of the SoTA Likelihood Ratio Attack (LiRA) from \cite{carlini2022membership} to the unlearning setting. The `online' version refers to the need to train new models for every membership query. Notably, they continue the precedent set in \cite{carlini2022membership} by showing Receiver Operating Characteristic (ROC) curves with log-scales to show performance at very low false-positive rates - better demonstrating worst-case outcomes. The effectiveness of LiRA as an MIA makes it a much more realistic estimate of privacy.

One downside of \cite{kurmanji2023towards}'s online-LiRA implementation is its computational complexity. The attack entails training 256 `shadow' models (each on a random half of the training set) and then running an unlearning algorithm 10,000 times on random forget sets for \textit{each} of the 256 shadow models. This amounts to over 2.5 million unlearning runs per algorithm. For any fine-tuning based algorithm, or any algorithm dealing with even moderately large models or data sets, this sort of evaluation is impractical. 

To remedy this gap, \textbf{we adapt the \textit{offline} version of LiRA for unlearning} in our benchmarking framework. `Offline' means that no new shadow models need to be trained for new membership queries. This setup only requires a single unlearning run, making this attack much more feasible - especially in the iterative unlearning setting (Section \ref{sec:iterative-unlearning}). While it is not as strong an attack as online-LiRA, and therefore worse for privacy estimation, it is much stronger than the basic Logistic Regression MIA, so we recommend its use in cases where the online-LiRA attack is computationally infeasible.

While MIAs are often just used to rank unlearning algorithms, some approaches have been made to more directly estimate DP privacy parameters ($\epsilon$, $\delta$). The 2023 NeurIPS Machine Unlearning Competition \cite{neurips-2023-machine-unlearning} based their metric on group-level DP, a stricter formulation of DP which considers datasets differing by up to $k$ examples, which makes sense for unlearning as requests are likely to be batched. In fact, the competition metric used estimates of worst-case MIA performance to estimate $\epsilon$ for \textit{each} example in $\mathcal{D}_f$, producing an entire distribution of privacy levels. While \cite{neurips-2023-machine-unlearning} averaged the per-example estimates of $\epsilon$ into a single score, \textbf{we re-implement the per-example $\epsilon$ estimation to preserve the full distribution of privacy levels}, which we find to be a valuable point of comparison.

\subsection{Update Leakage}

While intuitively one may think that unlearning can only result in positive outcomes with respect to the privacy of forgotten data, \cite{chen2021machine} demonstrate that, in regards to an attacker who already has the outputs of the base model, the act of unlearning may result in \textit{worse} privacy than not unlearning at all. The idea of model update-leakage - where an adversary uses the difference in behaviors of two models as an attack vector - has been previously studied in iterative learning scenarios \cite{salem2020updates, zanella2020analyzing}, and \cite{chen2021machine} demonstrate that the issue is still relevant in the unlearning setting. \cite{chen2021machine} find that update-leakages are often stronger than standard attacks for both fully retrained and unlearned models, although the effect is weaker for the latter. In this regard unlearning may actually be preferential to retraining in terms of privacy, as it could find an optimal middle ground between the two types of attacks. We therefore evaluate unlearning algorithms on update-leakage attacks to:
\begin{enumerate}
    \item Demonstrate an additional benefit of unlearning over retraining from scratch by showing less susceptibility to update-leakage attacks.
    \item Ensure a `do no harm' criteria is satisfied by showing an update-leakage attack does no worse than an attack on the base model. 
\end{enumerate}
To the best of our knowledge (through review of citations of \cite{chen2021machine}), the only unlearning algorithm benchmarked on an update-leakage attack, besides SISA in \cite{chen2021machine}, is GraphEraser \cite{chen2022graph}, an unlearning algorithm for graph data. \textbf{We implement and run evaluations on the update-leak attack from \cite{chen2021machine}}, which works very similarly to online-LiRA, except uses outputs from both the base and the unlearned models.

\subsection{Iterative Unlearning} \label{sec:iterative-unlearning}

The final piece of a comprehensive unlearning evaluation is studying how unlearning affects a model after repeated applications of an algorithm. If unlearning can truly be a replacement for full retraining, and deployers of ML models fully comply with data removal requests and destroy the base model trained on $\mathcal{D}_f$, then unlearning must be \textit{iteratively} applied to unlearned models. Few papers consider this set-up, and those that do are usually special cases where unlearning can be performed in closed form or with convex optimization \cite{gupta2021adaptive} \cite{neel2020descenttodelete} \cite{guo2019certified}. Surprisingly, we were able to locate only one unlearning paper that evaluates iterative test accuracy  \cite{chen2023unlearn}. We have not encountered an iterative test accuracy evaluation in the vision domain. 

In the iterative setting, it is required at each iteration to both ensure effective forgetting and, crucially, maintain model performance, as performance degradation tends to accumulate over time. Single forget set evaluations simply do not capture this important dimension of real-world performance. It should be noted that auditing privacy over many iterations may be prohibitively expensive, especially for attacks like online-LiRA that require hundreds of unlearning runs per iteration - so in this setting, our offline-LiRA implementation might be preferred. In either case, test-set performance can be computed directly after each unlearning iteration, so we advocate for its inclusion in any unlearning evaluation. \textbf{We implement an iterative unlearning pipeline} that handles the required dataset splitting and model management to allow for extending existing unlearning evaluations to the iterative setting -- we focus primarily on test-set accuracy.

\subsubsection{Preliminary Results} \label{prelim-results}

We conduct a preliminary evaluation of various unlearning algorithms on the CIFAR10 dataset \cite{krizhevsky2009learning} with a ResNet18 architecture \cite{he2016deep}, which was chosen due to its small computational footprint. We evaluate the following baselines and unlearning algorithms: Identity (baseline), Retrain (baseline), Finetune (baseline, finetune on $\mathcal{D}_r$), RandLabel \cite{graves2021amnesiac}, BadTeach \cite{chundawat2023can}, SCRUB+R \cite{kurmanji2023towards}, SSD \cite{foster2023fast}, and SSD+FT (SSD followed by Finetune).

For each unlearning iteration, $\mathcal{D}_f$ is constructed by sampling 1\% of $\mathcal{D}_{train}$, conditioned on samples that have not already been forgotten in prior iterations. The sequence of forget sets is identical across all experiments. The Identity algorithm is a control where no unlearning is applied. The base model used to test each unlearning algorithm is a ResNet18 model trained for 30 epochs. The hyperparameters for each unlearning algorithm were discovered by running 100 trials with the Optuna framework \cite{akiba2019optuna}, optimized for minimizing $|0.5 - \text{MIA Accuracy}|$ (aiming for the MIA to do no better than random) and maximizing validation accuracy.

\begin{figure}[!t]
\centering
\includegraphics[width=\linewidth]{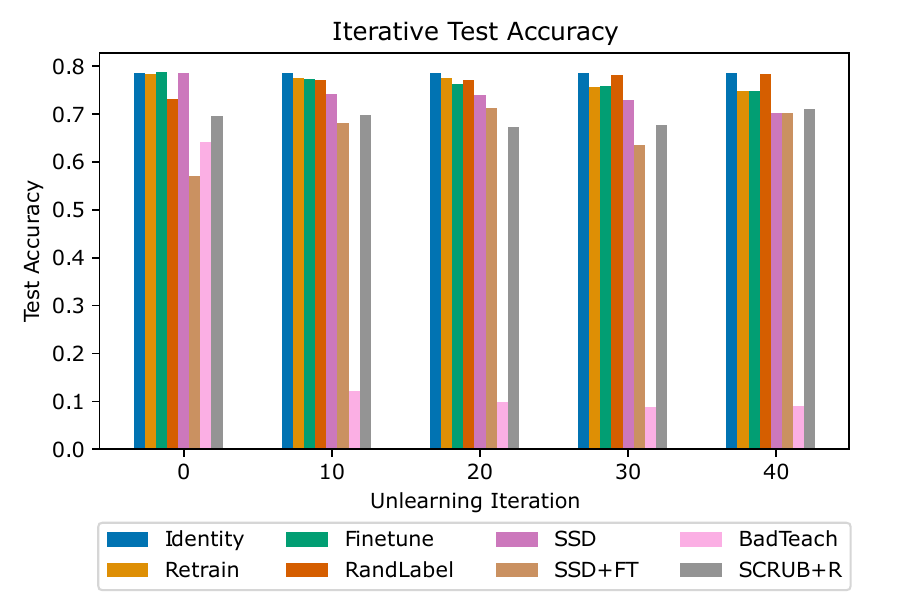}
\caption{Iterative unlearning results for ResNet18 on CIFAR10.}
\label{fig_sim}
\end{figure}

In this iterative setting we find notable discrepancies in test accuracy amongst the various algorithms. As shown in Figure \ref{fig_sim}, BadTeach rapidly degrades model performance to random guessing, while other algorithms are able to maintain or in some cases increase accuracy over time. 
While these results only analyze test-set accuracy, and do not characterize either privacy or runtime, they demonstrate an important facet that we have observed to be considered in only one other unlearning evaluation \cite{chen2023unlearn}.

\section{Discussion}

Overall, our initial tests show mixed results across unlearning algorithms, emphasizing the need for a holistic evaluation across all three major requirements (test accuracy, runtime, and privacy). For our privacy evaluations, we find that success in defending the basic Logistic Regression MIA, utilized in most unlearning evaluations, does \textit{not} necessarily translate to success under stronger MIAs. In terms of update-leakage attacks, we find that only some of the tested algorithms perform better than retraining, but none have yet been found to leak extra information. As we saw in Section \ref{prelim-results}, there are considerable discrepancies in test accuracy in the iterative setting. We also find that hyperparameter tuning is a key aspect of algorithm performance, and different hyperparameters trade off performance and privacy differently. If optimizing hyperparameters for privacy with an expensive attack like online-LiRA the tuning process can be extremely slow, potentially presenting a significant barrier to real-world adoption. 

\section{Future Work}

We will run evaluations on a variety of model architectures and datasets to provide a rigorous presentation of the state of the field. Our code base will be open sourced to provide an easy-to-use toolkit for designing and evaluating machine unlearning algorithms.

\ifCLASSOPTIONcompsoc
  \section*{Acknowledgments}
\else
  \section*{Acknowledgment}
\fi


Copyright 2024 Carnegie Mellon University. This material is based upon work funded and supported by the Department of Defense under Contract No. FA8702-15-D-0002 with Carnegie Mellon University for the operation of the Software Engineering Institute, a federally funded research and development center. The view, opinions, and/or findings contained in this material are those of the author(s) and should not be construed as an official Government position, policy, or decision, unless designated by other documentation. This material is licensed under a Creative Commons Attribution-NonCommercial-ShareAlike 4.0 International License.



\bibliographystyle{IEEEtran}
\bibliography{IEEEabrv,./bibliography}
%


\end{document}